\documentclass[letterpaper, 10 pt, conference]{ieeeconf}  % Comment this line out if you need a4paper

\IEEEoverridecommandlockouts                              % This command is only needed if 
\usepackage{graphics} % for pdf, bitmapped graphics files
\usepackage{amsmath} % assumes amsmath package installed
\usepackage{booktabs}

\usepackage{cite}

\usepackage{mwe} % to get dummy images
\usepackage{hyperref}

\title{\LARGE \bf
Deep Neural Encoder-Decoder Model to Relate fMRI Brain Activity with Naturalistic Stimuli
}

\author{Florian David$^{1,}$\textsuperscript{\textsection}, Michael Chan$^{1,2,}$\textsuperscript{\textsection}, Elenor Morgenroth$^{1,2,3}$, Patrik Vuilleumier$^{3,4}$, Dimitri Van De Ville$^{1,2}$% <-this % stops a space
\thanks{\textsuperscript{\textsection}Equal contribution}% <-this % stops a space
\thanks{$^{1}$Neuro-X Institute, Ecole Polytechnique Fédérale de Lausanne (EPFL), Geneva, Switzerland}%
\thanks{$^{2}$Department of Radiology and Medical Informatics, University of Geneva, Geneva, Switzerland}%
\thanks{$^{3}$Swiss Center for Affective Sciences, University of Geneva, Geneva, Switzerland}%
\thanks{$^{4}$Geneva Neuroscience Center, University of Geneva, Geneva, Switzerland}%
}

\begin{document}

\maketitle
\thispagestyle{empty}
\pagestyle{empty}

%%%%%%%%%%%%%%%%%%%%%%%%%%%%%%%%%%%%%%%%%%%%%%%%%%%%%%%%%%%%%%%%%%%%
\begin{abstract}

We propose an end-to-end deep neural encoder-decoder model to encode and decode brain activity in response to naturalistic stimuli using functional magnetic resonance imaging (fMRI) data. Leveraging temporally correlated input from consecutive film frames, we employ temporal convolutional layers in our architecture, which effectively allows to bridge the temporal resolution gap between natural movie stimuli and fMRI acquisitions. Our model predicts activity of voxels in and around the visual cortex and performs reconstruction of corresponding visual inputs from neural activity. Finally, we investigate brain regions contributing to visual decoding through saliency maps. We find that the most contributing regions are the middle occipital area, the fusiform area, and the calcarine, respectively employed in shape perception, complex recognition (in particular face perception), and basic visual features such as edges and contrasts. These functions being strongly solicited are in line with the decoder's capability to reconstruct edges, faces, and contrasts. All in all, this suggests the possibility to probe our understanding of visual processing in films using as a proxy the behaviour of deep learning models such as the one proposed in this paper.
\end{abstract}

\vspace{5px}


%%%%%%%%%%%%%%%%%%%%%%%%%%%%%%%%%%%%%%%%%%%%%%%%%%%%%%%%%%%%%%%%%%%%%%%%%%%%%%%%
\section{Introduction}
\label{sec:intro}

The study of brain processes during visual stimuli is a rapidly advancing field with significant implications for brain-computer interfaces (BCI) and cognitive neurosciences. Predicting the brain's response to visual input (encoding) and reconstructing perceived visual stimuli based on brain activity (decoding) are key challenges, especially when working with stimuli that closely mirror everyday experiences. Early studies made considerable progress in encoding and decoding of static images \cite{Takagi2022, Shen2019, Beliy2019, Gaziv2022, Simonyan2014}, contributing significantly to our understanding of visual information representations in the brain. Moreover, in an effort to provide a more accurate representation of real-world visual experiences, much of the recent work has used movies as stimuli \cite{Nishimoto2011, Han2019, Khosla2021, Wen2017, Kupershmidt2022, Le2022}. 

Many attempts to decode brain activity involve generative models, such as diffusion models \cite{Takagi2022}, generative adversarial networks (GANs) \cite{Shen2019} or variational autoencoders (VAEs) \cite{Han2019}. While these models generate highly realistic images by learning latent structures of fMRI data, they often suffer from hallucinations and sometimes fail to faithfully represent the actual stimuli, as pointed out by Shirakawa et al. \cite{Shirakawa2024}. A sound alternative is the use of convolutional neural networks (CNNs) \cite{Wen2017, Kupershmidt2022, Le2022}, which are less prone to hallucinations and may offer better zero-shot predictions on unseen visual stimuli. CNNs are well-suited to extract visual features from images and videos that align with neural processing in the human visual cortex \cite{Yamins2016}, but they often rely on large paired datasets of stimuli and corresponding brain recordings, which are scarce, especially for natural movies. Recent approaches, such as self-supervised models \cite{Beliy2019, Gaziv2022, Kupershmidt2022}, address this limitation by reducing the need for paired samples. 

Natural movies present several unique challenges for brain encoding and decoding due to their dynamic nature and high temporal resolution (around 30 frames per second). In contrast, fMRI recordings have a coarse temporal resolution with sample rates of 1.3 to 2 seconds (about 0.5 Hz), creating a temporal discrepancy that limits the capture of rapid neural dynamics. Additionally, the blood oxygenation level dependent (BOLD) signal, recorded in fMRI to localize brain activity, reflects slow and delayed hemodynamic changes and introduces temporal blurring that makes decoding fast-changing visual content more difficult. Thus, advanced models are required to bridge this temporal gap.

We introduce a novel end-to-end encoder-decoder model that predicts fMRI recordings from natural videos and reconstructs movie frames from voxel activations. Our main contributions are as follows: (1) While many studies focus on decoding, we emphasize voxel prediction, and also use the decoder as a regularizer to enhance encoding performances. (2) We employ a rich dataset of free-viewing naturalistic stimuli to closely mimic everyday life visual experiences and capture a valid representation of brain activity. (3) By using a rigorous dataset-splitting method, we demonstrate that our model captures temporal and spatial information effectively on both familiar and completely novel visual content, indicating its robustness. (4) We provide insights into the mechanisms underlying visual reconstruction through saliency map analysis, highlighting brain regions most influential in decoding, which turn out to be the same regions necessary for the decoded features in the representation of visual stimuli for the human brain. 

\begin{figure*}[!t]
    \begin{minipage}[t]{0.81\textwidth}
        \centering
        \centerline{\includegraphics[width=\textwidth]{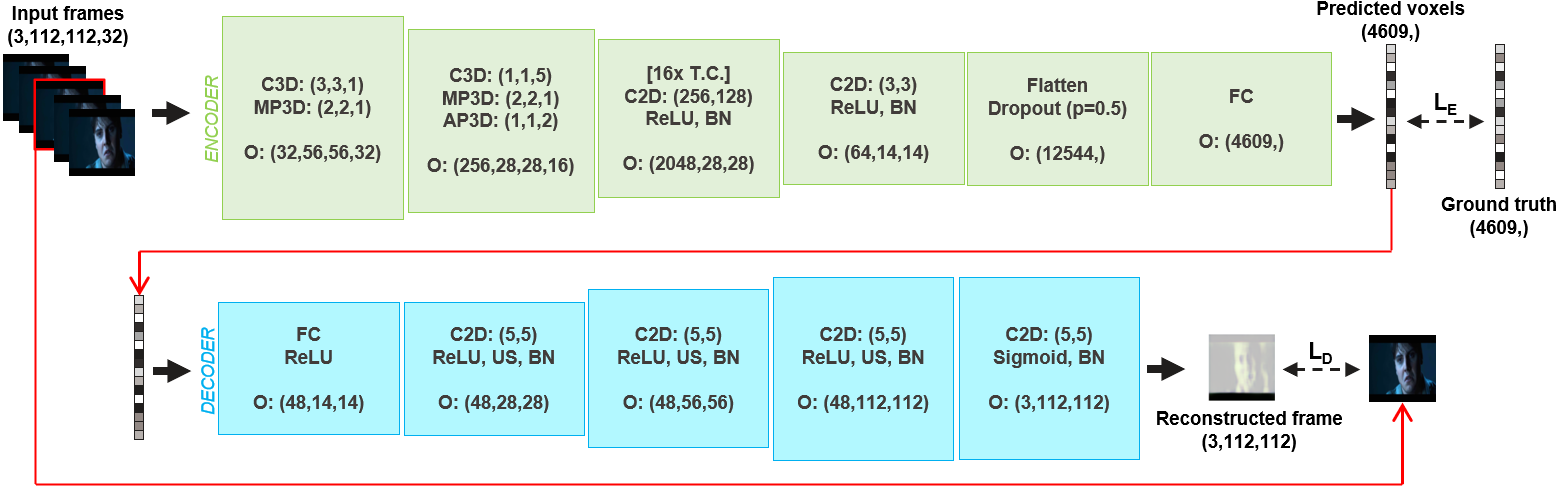}}
        \centerline{(a) Model architecture.}
    \end{minipage}%
    \hspace{1mm}
    \begin{minipage}[t]{0.17\textwidth}
        \centering
        \centerline{\includegraphics[width=\textwidth]{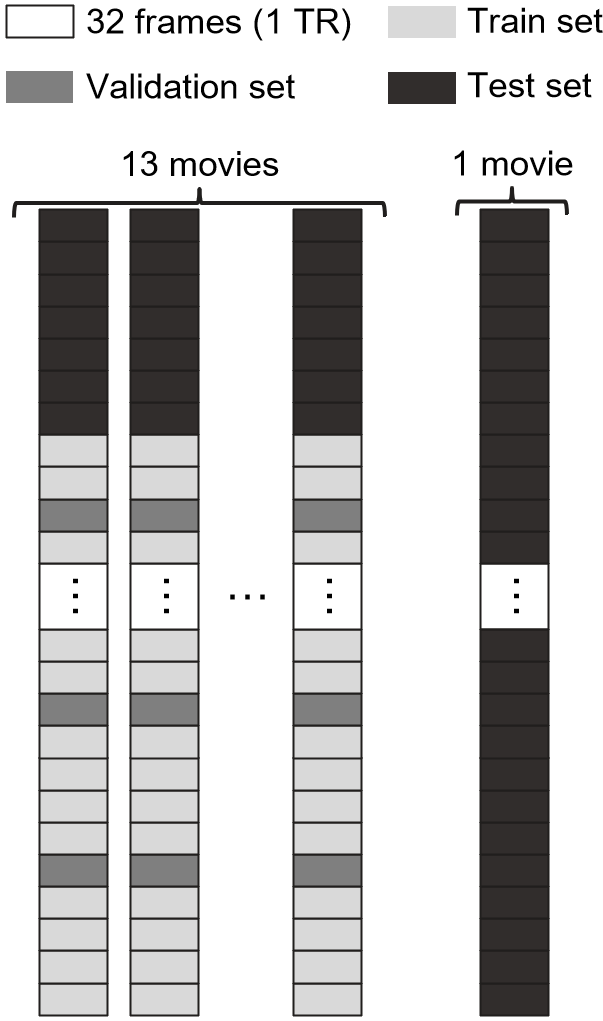}}
        \centerline{(b) Splitting method.}
    \end{minipage}%
    \hfill
    \caption{Overview of the method. \textbf{(a)} The encoder-decoder model consists of two distinct CNNs trained end-to-end with various convolutional ($C3D$, $C2D$), pooling ($MP$, $AP$), and fully connected ($FC$) layers. Transformations include $ReLU$, batch norm ($BN$), dropout, upsampling ($US$), and sigmoid. $T.C.$ refers to temporal combinations, and $O$ refers to the output shape of the different layers. $L_{E}$ and $L_{D}$ represent encoder and decoder losses. \textbf{(b)} The dataset was preprocessed such that 32 consecutive frames correspond to 1 fMRI volume (1 TR), averaged across all 30 subjects. The dataset was split into training, validation, and test sets. The test set includes a full movie (\textit{*13-YA}) and the last 20\% of volumes from the remaining 13 movies.} 
    \label{fig:split_architecture}
\end{figure*}

\section{Methods}
\label{sec:Methods}

\subsection{Data preprocessing}
\label{sec:data_preprocessing}

In our work, we use the Emo-FilM dataset~\cite{morgenroth2024}, which contains fMRI recordings of 30 subjects watching 14 short films. The acquisition, preprocessing, and quality of the data are described in \cite{morgenroth2024}. We used minimally preprocessed fMRI recordings registered to the MNI brain template. For simplicity, in the following sections, we use the fMRI repetition time (TR) as temporal unit, where 1 TR corresponds to 1.3 seconds, with film runs ranging from 309 TRs to 790 TRs in length. The films cover a wide range of visual stimuli, and subjects were allowed to explore the visual field while watching. 

We applied additional preprocessing to the dataset. Notably, we removed each voxel’s baseline signal and normalized the time series to ensure uniformity across the dataset. We introduced a delay of 4 TRs between fMRI and film to account for hemodynamic response. 
We used the Schaefer 1000 parcellation \cite{Schaefer2017} to produce a mask of relevant regions associated with visual functions, specifically targeting the striate and extrastriate cortical areas. This mask reduced our data to $15364$ voxels. Similarly to \cite{Kupershmidt2022}, we further refined the parameter space of our models by selecting the top 30\% voxels with the highest signal-to-noise ratio (SNR). This resulted in $4609$ remaining voxels. SNR was computed by dividing the signal variance (based on the mean activation across subjects for each time point and voxel) by the noise variance (based on the mean activation across time for each subject and voxel).
%This resulted in $4609$ remaining voxels. SNR was computed by dividing the mean activation across subjects for each time point and voxel by the mean activation across time for each subject and voxel. 
In a final preprocessing step, we averaged the fMRI signal across subjects, mitigating subject-specific variability and emphasizing shared patterns. 
We also applied preprocessing to the films' frames to maintain spatial and temporal consistency. All films were standardized to $112\times 112$ pixels at 32 frames per TR, uniformly up- or down-sampled as necessary. We exclusively considered the visual and divided the films into N chunks of 32 RGB frames. Note that during the fMRI acquisition, sound was available.

% The SNR was first computed for individual movies across the different subjects. We computed the 'signal' as the mean activation across subjects for each time point and voxel, and derived the signal variance as the variance of these signal means across time for each voxel. Then, we computed the 'noise' as the mean activation across time for each subject and voxel, and derived the noise variance as the variance of these noise means across subjects for each voxel. The SNR of each voxel was then obtained by dividing the signal variance by the noise variance, resulting in an SNR value per voxel and per movie. By averaging these SNR values across all movies, we established a criterion to rank and select voxels demonstrating high signal fidelity relative to their noise level.

\subsection{Models}
\label{sec:models}

The model in this study consists of two distinct CNNs, an encoder (E) and a decoder (D), which can function independently. The architectures of the sub-models were inspired by the work of \cite{Kupershmidt2022} (Fig.~\ref{fig:split_architecture}a).

The encoder predicts fMRI activity from movie frames. It is trained by minimizing the mean squared error (MSE) while maximizing cosine similarity between the predicted voxels and their ground truth (across time). The encoder loss function is defined as:
\begin{equation}
    L_{E}({\bf v, \hat{v}}) = \text{MSE}({\bf v, \hat{v}}) + \alpha \cdot \cos(\angle({\bf v, \hat{v}})),
    \label{loss_encoder}
\end{equation}
where $\bf v$ and $\hat{\bf v}$ are the ground truth fMRI recordings (voxels) and the encoder predictions, respectively, and $\cos(\angle(\cdot, \cdot))$ represents the cosine distance between vectors. The hyperparameter $\alpha$ is set to $0.5$.

The decoder reconstructs movie frames from fMRI data. It is trained such that the reconstructed frames are both visually and structurally consistent with the target frames. The decoder loss function is defined as:
\begin{equation}
    L_{D}({\bf f, \hat{f}}) = \beta \cdot L_{psim}({\bf f, \hat{f}}) + \gamma \cdot L_{ssim}({\bf f, \hat{f}}) + \delta \cdot L_{tv}({\bf \hat{f}}),
    \label{loss_decoder}
\end{equation}
where $\bf f$ and $\bf \hat{f}$ are the target and reconstructed frames, respectively. While the encoder takes as input chunks of 32 frames, we define the target frames of the decoder as the middle frame of each chunk. The loss combines a perceptual similarity metric $L_{psim}$ \cite{Gaziv2022} based on the differences between the five feature-extractor layers of the pretrained VGG16 model \cite{Simonyan2014}, a structural similarity (SSIM) metric $L_{ssim}$ based on texture, luminance, and contrast, and a total variation regularization term $L_{tv}$ based on smoothness of the reconstruction. Hyperparameters are set as $\beta = 0.35$, $\gamma = 0.35$, and $\delta = 0.30$ \cite{Kupershmidt2022}.

The end-to-end encoder-decoder model takes both movie frames and ground truth fMRI as input, with a combined loss function defined as:
\begin{equation}
    L_{ED}({\bf v, \hat{v}, f, \hat{f}}) = \epsilon \cdot L_{E}( {\bf v, \hat{v}}) + (1-\epsilon) \cdot L_{D}( {\bf f, \hat{f}} ),
\end{equation}
where $\epsilon$ is a hyperparameter ($\epsilon = 0.5$). In this setup, the decoder reconstructs movie frames using the fMRI predicted by the encoder, serving as an additional regularization module to enhance encoding performance.

\subsection{Models validation}

To train the model efficiently, we split the dataset as follows: for 13 out of the 14 movies, the first 80\% of the frames were uniformly divided into training and validation sets (4:1 ratio), while the remaining 20\% of frames and the complete 14th movie (\textit{YouAgain}, \textit{*13-YA}) constituted the test set (see Fig.~\ref{fig:split_architecture}b). This partitioning ensures comprehensive model evaluation and generalizability. With regard to fMRI time courses, training, validation, and testing are carried out on subject-averaged fMRI corresponding to the selected movie frames, yielding a group model per training. 

\begin{figure}[!b]
        \centering
        \centerline{\includegraphics[width=\linewidth]{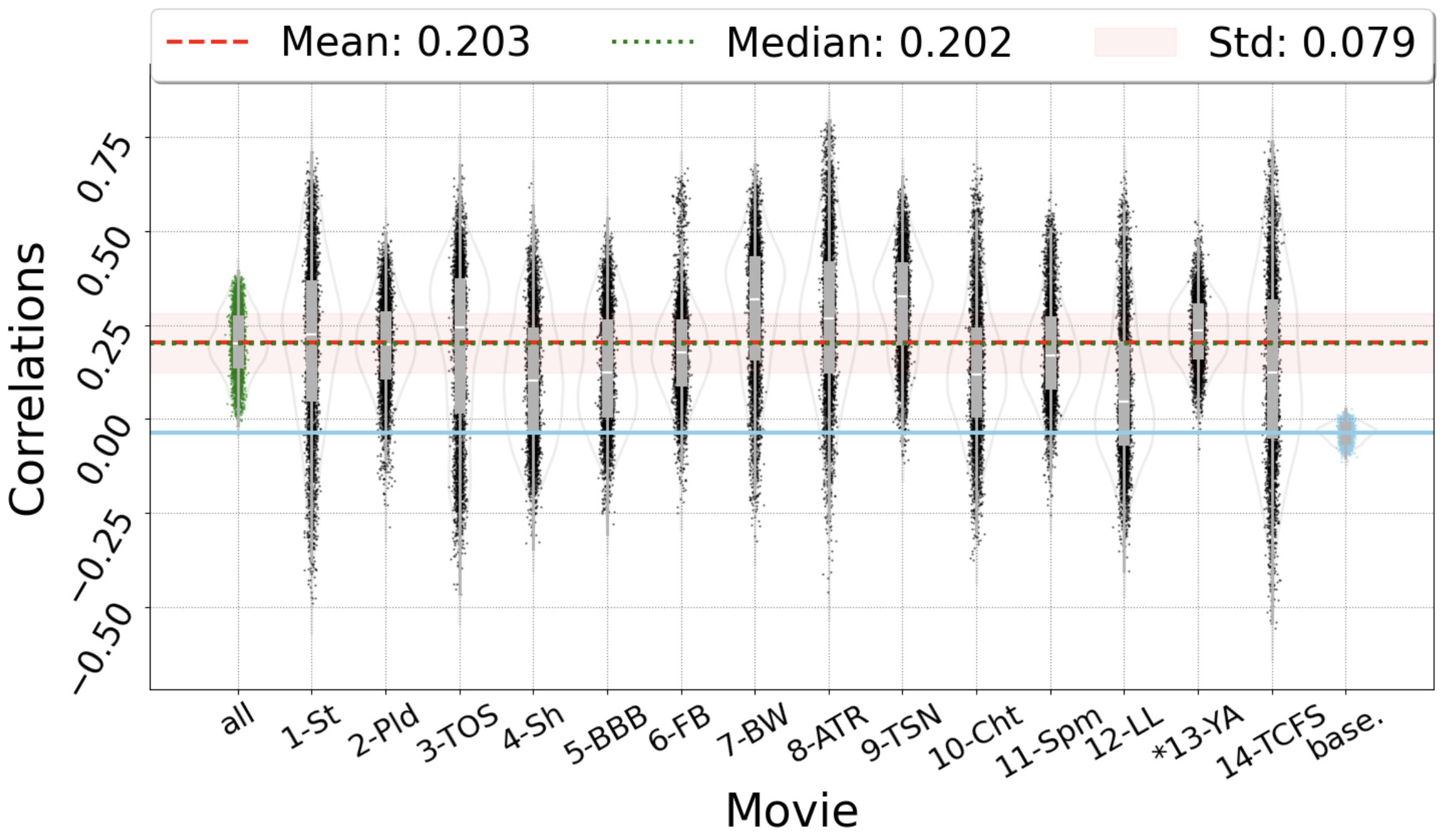}}
    \caption{Encoder performance. Pearson correlations between true and encoder-predicted voxel activation over time. Black dots represent correlations across consecutive test TRs from specific movies; green dots represent correlations across all concatenated test TRs; cyan dots and horizontal line represent correlations and mean correlations against shuffled time courses, serving as null distribution. The encoder achieved an average correlation score of 0.203 with an average MSE of 1.594 ($N=4609$ voxels).}\label{fig:encoder_performance}
\end{figure}

\begin{figure}[!b]
        \centering
        \centerline{\includegraphics[width=\linewidth]{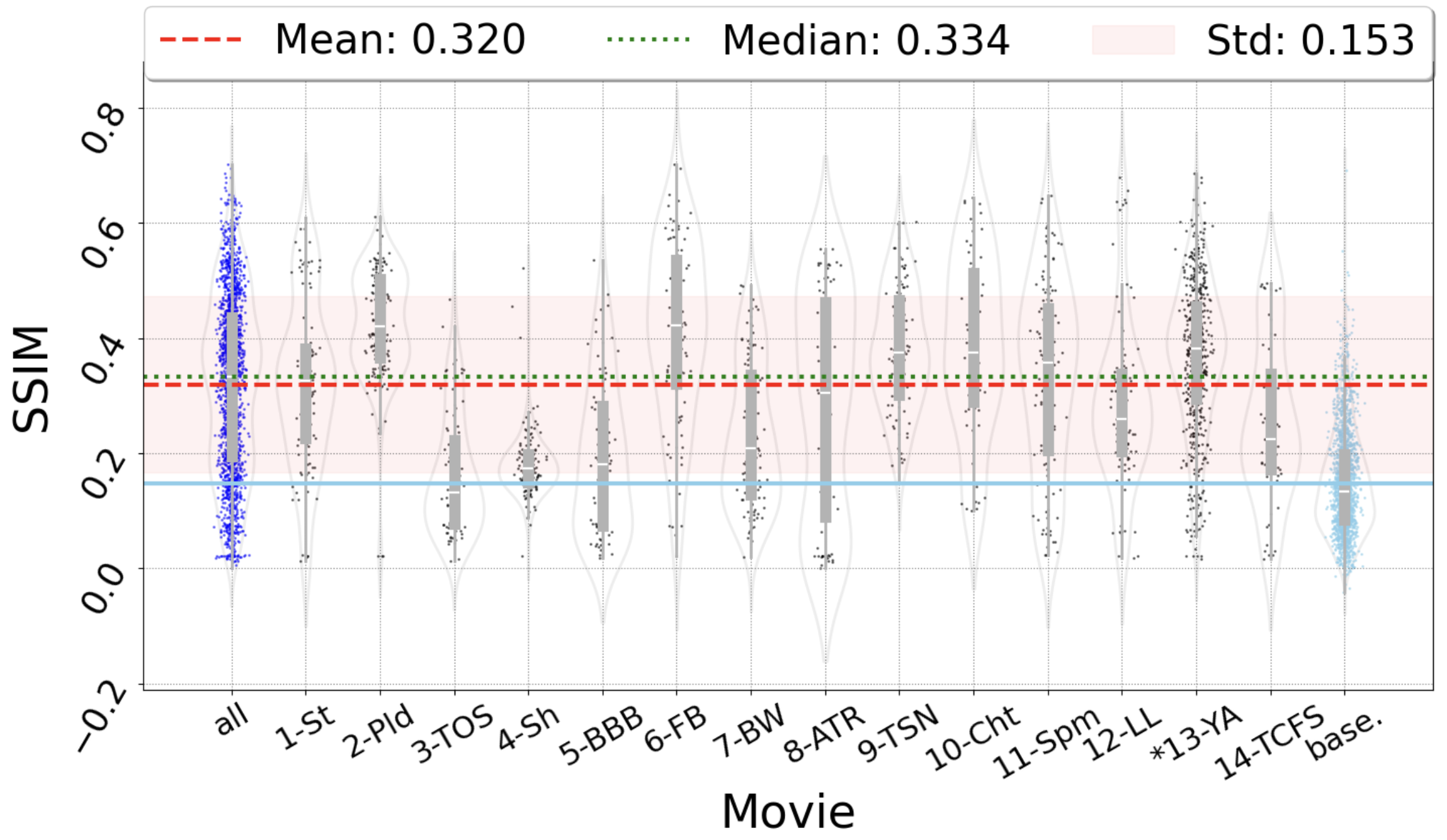}}
    \caption{Decoder performance. SSIM scores between true and decoder-predicted movie frames. Black dots represent SSIM values across test frames from specific movies; blue dots represent SSIM values across all test frames; cyan dots and horizontal line represent SSIM scores and mean SSIM scores against shuffled movie frames, serving as null distribution. The decoder achieved an average SSIM score of 0.320 with an average MSE of 0.172 ($N=1971$ frames).}\label{fig:decoder_performance}
\end{figure}

\section{Results}
\label{sec:results}

% We present the results obtained with the best encoder-decoder model, referred to as E-D ($4609$) in Table \ref{tab:model_comparison} being the encoder-decoder model computed on a mask of 4609 voxels defined in \ref{sec:data_preprocessing}. Its hyperparameters were set as described in Section \ref{sec:models}, and it was trained using the Adam optimizer at a learning rate of $10^{-4}$ for 11 epochs. For better readability, the names of the 14 movies were abbreviated, with \textit{*13-YA} corresponding to the movie \textit{YouAgain} which was excluded from training.

As mentioned in Section \ref{sec:models}, we set values of the hyperparameters $\alpha$, $\beta$, $\gamma$, and $\delta$ following the work of Kupershmidt et al. \cite{Kupershmidt2022}, and then performed a grid search exclusively for $\epsilon \in [0, 1]$ and the number of training epochs in $[5, 30]$. In this section, we present the results obtained with the best encoder-decoder model, referred to as E-D ($4609$) in Table \ref{tab:model_comparison}, where the number $4609$ refers to the voxel mask used and described in Section \ref{sec:data_preprocessing}. It was trained using the Adam optimizer at a learning rate of $10^{-4}$ for 11 epochs with $\epsilon = 0.5$. For better readability, the names of the 14 movies were abbreviated, with \textit{*13-YA} corresponding to the movie \textit{YouAgain} which was excluded from training.

\subsection{Encoding performance}
\label{sec:encoding_performance}

To evaluate the encoder's performance, we computed the Pearson correlation coefficients and Mean Squared Error (MSE) between the ground truth and the predicted voxel activations. As shown in Fig. \ref{fig:encoder_performance}, the model successfully captured the temporal dynamics of the fMRI data, and results were consistent across the different movie segments of the test set.

\subsection{Decoding performance}
\label{sec:decoding_performance}

\begin{figure}[!b]
    \centering
    \centerline{\includegraphics[width=\linewidth]{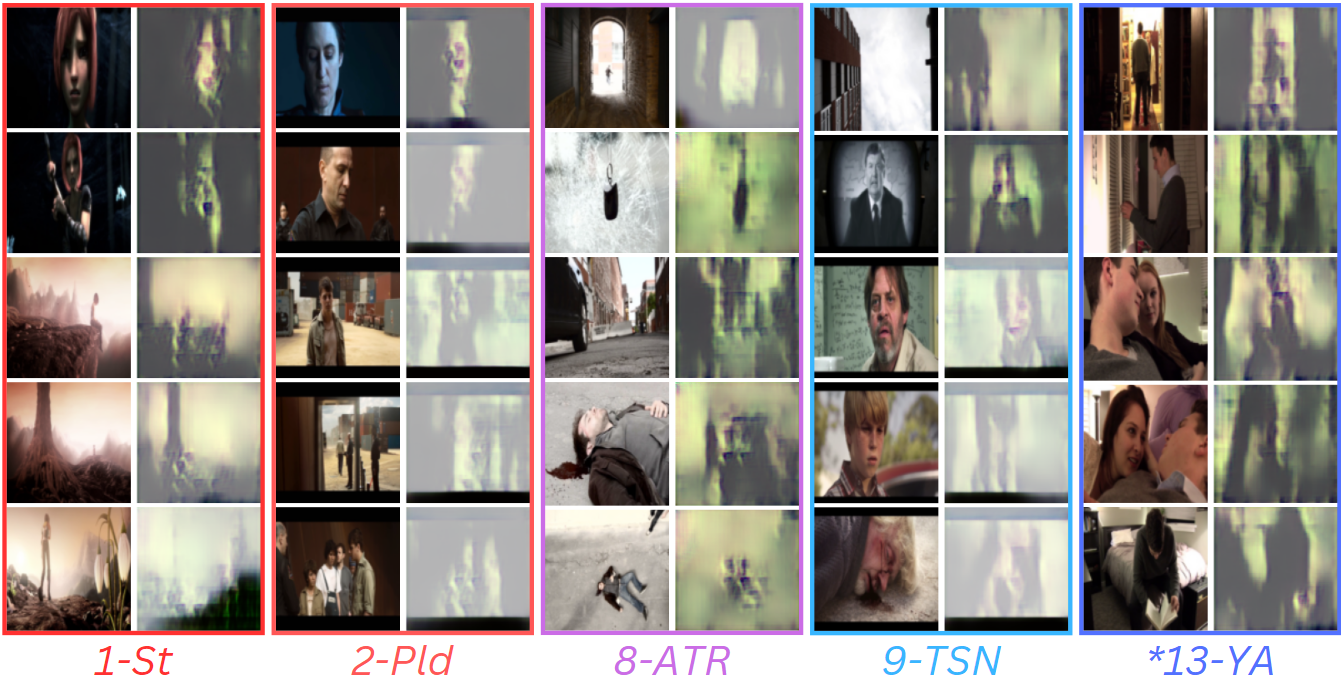}}
    \caption{Examples of reconstructed frames by the decoder across the test set. Each pair shows the ground truth frame (left) and the decoder's reconstruction (right).}
    \label{fig:reconstructions}
\end{figure}

To assess the decoder's ability to reconstruct visual stimuli from the fMRI data predicted by the encoder, we computed the SSIM and MSE between the ground truth target frames and the ones predicted by the decoder. Fig. \ref{fig:decoder_performance} shows that the decoder consistently captured visual details from neural activations across the test set, with examples of reconstruction provided in Fig. \ref{fig:reconstructions}.

\begin{table}[!b]
\centering
\small % Reduce font size if needed
\begin{tabular}{l c c c c}
\toprule & E-D (4609) & E-D (15364) & E (4609) & E (15364) \\
\midrule
$E_{\text{corr}} \uparrow$ & $\textbf{0.202}^{*,**}$ & $0.140^{*}$ & $0.139^{**}$ & $0.108$ \\
$E_{\text{mse}} \downarrow$ & $\textbf{1.595}$ & $1.721$ & $1.722$ & $1.783$ \\
$D_{\text{ssim}} \uparrow$ & $\textbf{0.334}^{***}$ & $0.330^{***}$ & / & / \\
$D_{\text{mse}} \downarrow$ & $0.162$ & $\textbf{0.072}$ & / & / \\
\bottomrule
\end{tabular}
\caption{Comparison of median model performances based on architecture and visual mask used. The model is either the end-to-end encoder-decoder (E-D) or the encoder alone (E), and the mask covers either all the visual voxels (15364) or only the ones with highest SNR (4609) as described in Section \ref{sec:data_preprocessing}. \textit{E-D (4609)} corresponds to the best model. Mann-Whitney U tests on metric distributions: *($n1 \times n2=4609\times15364$; $p<0.001$; $\delta_{c}=0.392$), **($n1 \times n2=4609\times4609$; $p<0.001$; $\delta_{c}=0.383$), ***($n1 \times n2=1971\times1971$; $p=0.293$; $\delta_{c}=0.019$).}

\label{tab:model_comparison}
\end{table}

We further explored the impact of the decoder and visual mask on encoder performance. Table \ref{tab:model_comparison} summarizes the median performance metrics across four configurations, i.e., encoder-decoder (E-D) on masks of 4609 and 15364 voxels and similarly for the encoder alone (E). The first column corresponds to the best model described in the rest of this study. Given that the metrics are not normally distributed and sample sizes were large, we performed non-parametric Mann-Whitney U tests and calculated Cliff's delta ($\delta_{c}$) for comparison of model performances. We found statistically significant improvement in encoding performances when using the encoder-decoder model with the $4609$-voxels mask compared to using the encoder alone or training over all $15364$ voxels.

\begin{figure}[!b]
        \centering
        \centerline{\includegraphics[width=\linewidth]{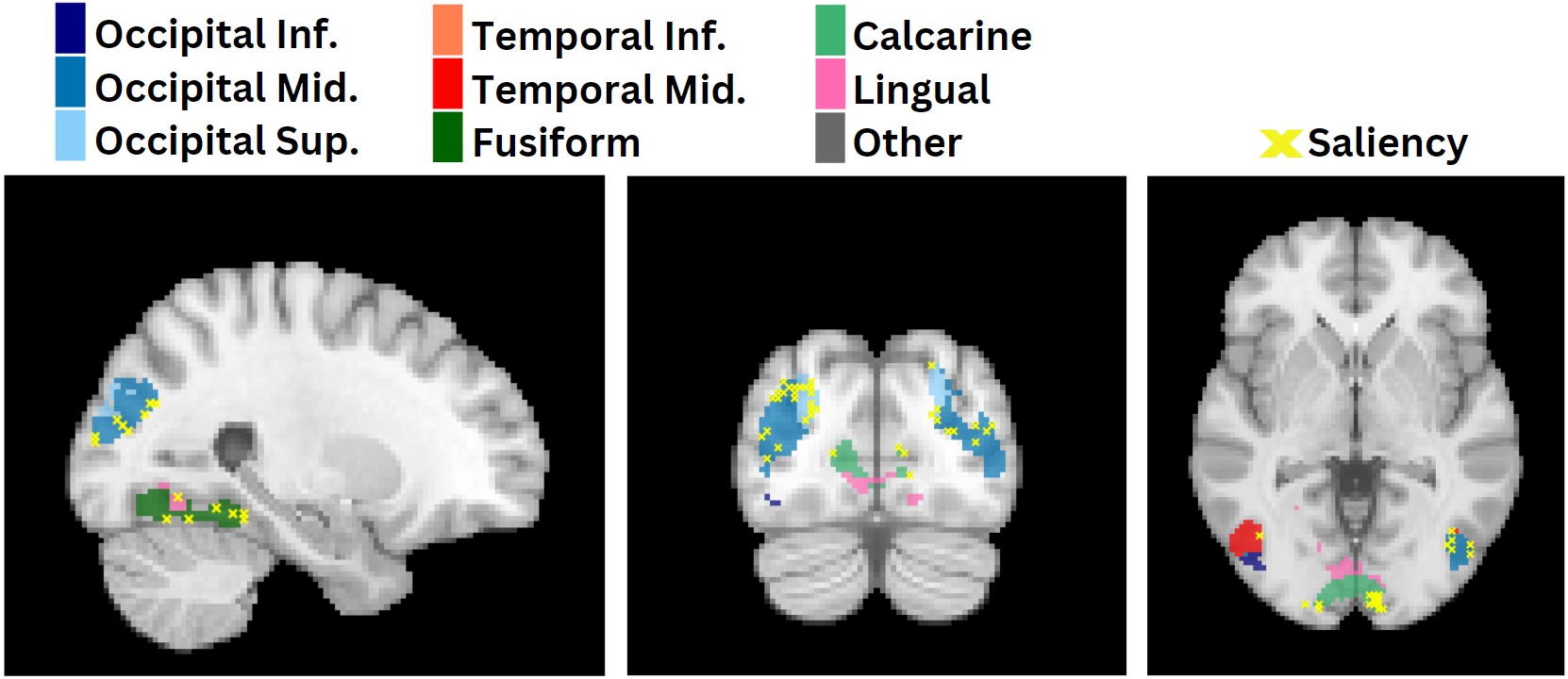}}
    \caption{Main contributing brain regions to movie frame decoding found by selecting the top $20\%$ saliency values (shown in yellow crosses). All colored areas are part of the selected voxels described in Section \ref{sec:data_preprocessing}. Anatomical regions were labelled from Automated Anatomical Labeling atlas 3 (AAL) \cite{Rolls2020}.}
    \label{fig:saliency}
\end{figure}

Finally, we generated saliency maps to identify voxels most influential in frame reconstruction. Saliency was computed by backpropagating SSIM gradients through the decoder, reflecting the voxel sensitivity of SSIM scores. Each voxel's saliency score was computed by summing the absolute saliencies across all reconstructions. A high saliency value indicates a strong influence of the voxel value on the decoding performance. We show 
in Fig.~\ref{fig:saliency} the topmost $20\%$ contributing voxels to frame reconstruction and observe that the relevant voxels are for the most part situated in the middle-superior occipital area, fusiform, and calcarine. We confirm this observation quantitatively by looking at contributions across regions in Table~\ref{tab:ratio_saliency}.

% To gain insight into the contribution of the different brain regions to the frame reconstructions, we computed saliency maps and identified the voxels carrying the most relevant information. The saliency maps were computed by backpropagating through the decoder the gradients of the SSIM values between the predicted movie frames and the corresponding ground truths with respect to the input (fMRI predicted by the encoder). These gradients reflect the sensitivity of the SSIM score to each feature of the input, indicating the magnitude of the influence of each voxel on the reconstruction score. For each voxel, we computed its saliency score by summing the absolute saliencies obtained for each reconstruction. A high saliency value indicates a strong influence of the voxel value on the decoding performance, whether it is positively or negatively (see Fig.~\ref{fig:saliency}).

\begin{table}[!b]
  \centering
  \small
  \begin{minipage}[t]{0.49\columnwidth}
    \begin{tabular}{lccc}
      \toprule
      (\%) & $\tfrac{A_{i}}{\Sigma A_{j}}$
 & $\tfrac{B_{i}}{\Sigma B_{j}}$ & $\tfrac{B_{i}}{A_{i}}$ \\
      \midrule
      Oc.I. & 4.5 & 8.4 & 37.7 \\
      \textbf{Oc.M.} & 32.1 & \textbf{27.5} & 17.1 \\
      Oc.S. & 7.6 & 12.3 & 32.5 \\
      Temp. & 13.8 & 8.5 & 33.9 \\
      \bottomrule
    \end{tabular}
  \end{minipage}%
  \hfill
  \begin{minipage}[t]{0.49\columnwidth}
    \begin{tabular}{lccc}
      \toprule
      (\%) & $\tfrac{A_{i}}{\Sigma A_{j}}$
 & $\tfrac{B_{i}}{\Sigma B_{j}}$ & $\tfrac{B_{i}}{A_{i}}$\\
      \midrule
      \textbf{Fusi.} & 15.9 & \textbf{12.6} & 15.8 \\
      \textbf{Calc.} & 13.4 & \textbf{20.9} & 31.3 \\
      Ling. & 11.5 & 6.8 & 11.8 \\
      Other & 1.4 & 3.0 & 42.2 \\
      \bottomrule
    \end{tabular}
  \end{minipage}
%\caption{Brain region contributions to frame decoding after selecting the top $20\%$ saliency values within the mask, as shown in Fig. \ref{fig:saliency}. $\tfrac{A_{i}}{\Sigma A_{j}}$ is the proportion of voxels that are located in region $i$. $\tfrac{B_{i}}{\Sigma B_{j}}$ is the proportion of threshold-exceeding saliency values that are in region $i$. $\tfrac{B_{i}}{A_{i}}$ is the proportion of voxels in region $i$ that have saliency values above the threshold.}
\caption{Brain region contributions to frame decoding after selecting the top $20\%$ saliency values within the mask, as shown in Fig. \ref{fig:saliency}. $A_{i}$ is the number of voxels that are located in region $i$. $B_{i}$ is the number of threshold-exceeding saliency values that are in region $i$.}
\label{tab:ratio_saliency}
\end{table}

% \subsection{Relating saliency and visual stimuli}
%  \begin{itemize}
%      \item human faces
%      \item low frequency bands
%      \item motions
%  \end{itemize}
\section{Discussion}
\label{sec:discussion}

In our paper, we present an encoding and decoding model leveraging the temporally correlated aspects of consecutive frames in movies, in contrast to the VAE perspective \cite{Han2019} through temporal CNNs. We first look into encoding performance to reveal the statistical significance of our models in predicting fMRI and then show better performance of the encoder-decoder in comparison to the encoder alone. We subsequently find that our model shows high decoder performance as the reconstructed images are qualitatively and quantitatively similar to the original ones. We obtain decoding results comparable to state-of-the-art reconstructions \cite{Kupershmidt2022, Le2022}. Our model shows generalizability both in fMRI prediction and movie reconstruction across segments unseen by the model.

Furthermore, saliency maps of our decoder highlight regions of the middle occipital area related to shape perception \cite{goodale1992separate}, the fusiform area, including that of the fusiform face area associated with face recognition \cite{kanwisher2006fusiform}, and the calcarine connected to primary visual features such as edges and contrasts \cite{Meadows2011}. This suggests that the decoder effectively performs fMRI readout following a representation akin to that of the human brain, leveraging the above-mentioned regions to perform image reconstruction. This also dissociates our work from potential semantic learning of generative models \cite{Shirakawa2024}. 

While we rigorously test model performance through various splitting methods, the learned weights may not perfectly predict other movie-fMRI datasets that have different fMRI preprocessing. This is explained by the nature of the data involved. As in the field, very different preprocessings of the fMRI exist and are known to yield different predictions \cite{sotero2023examining}. Despite this, it remains that the purpose of such a model is to be used as a way to study human brain visual representation within a given population. This work is motivated by the idea of employing a deep model to investigate brain function representation, as previously done in \cite{hebart2020revealing, gjolbye2023concept, brenner2024concept}.

Given the motive of the model, an interesting follow-up work would be to investigate behaviour of the decoding process under local perturbation of the fMRI volumes, e.g., interfering with activity in the fusiform and observing whether the decoder becomes unable to reconstruct certain patterns. 
A second axis of studies could include the improvement of the decoder and indirectly of the encoder by incorporating eye-tracking data or a saliency model of visual attention \cite{novin2023improved}, subsequently allowing for study of subject-level movie-fMRI volumes. 
Beyond saliency maps, which mainly highlight features with large effects on the decoding \cite{NEURIPS2018_294a8ed2}, a final future line of work could explore the use of Concept Activation Vectors (CAVs) \cite{kim2017interpretability} to enhance the interpretability of our encoder-decoder model and complement our saliency analysis. CAVs allow for the quantification of how human-interpretable concepts influence model activations by learning directional representations in their latent space. They could help identify which visual properties (e.g., contours, faces, motion) are most relevant for predicting voxel activity in the encoder, and which brain regions contribute to reconstructing visual features in the decoder.

\section{Compliance with Ethical Standards}
Conflict of Interest: The authors declare that they have no conflicts of interest. 

Funding: This work was supported by both the Swiss National Science Foundation Sinergia grant 180319 ``Charting emotion components and dynamics in the human brain using virtual reality and cinema'' and grant 209470 ``Precision mapping of electrical brain network dynamics with application to epilepsy''. 

This research study was conducted retrospectively using human subject data made available in open access by \href{https://openneuro.org/datasets/ds004892/versions/1.0.0}{OpenNeuro}. Ethical approval was not required, as confirmed by the license attached with the open access data.

%%%%%%%%%%%%%%%%%%%%%%%%%%%%%%%%%%%%%%%%%%%%%%%%%%%%%%%%%%%%%%%%%%%%%%%%%%%%%%%%

%%%%%%%%%%%%%%%%%%%%%%%%%%%%%%%%%%%%%%%%%%%%%%%%%%%%%%%%%%%%%%%%%%%%%%%%%%%%%%%%

%%%%%%%%%%%%%%%%%%%%%%%%%%%%%%%%%%%%%%%%%%%%%%%%%%%%%%%%%%%%%%%%%%%%%%%%%%%%%%%%
% \section*{APPENDIX}

% Appendixes should appear before the acknowledgment.

% \section*{ACKNOWLEDGMENT}

% The preferred spelling of the word ÒacknowledgmentÓ in America is without an ÒeÓ after the ÒgÓ. Avoid the stilted expression, ÒOne of us (R. B. G.) thanks . . .Ó  Instead, try ÒR. B. G. thanksÓ. Put sponsor acknowledgments in the unnumbered footnote on the first page.

%%%%%%%%%%%%%%%%%%%%%%%%%%%%%%%%%%%%%%%%%%%%%%%%%%%%%%%%%%%%%%%%%%%%%%%%%%%%%%%%

% References are important to the reader; therefore, each citation must be complete and correct. If at all possible, references should be commonly available publications.

% \begin{thebibliography}{99}

% \end{thebibliography}

\bibliographystyle{ieeetr}

\bibliography{refs}

\end{document}